\documentclass[10pt, journal]{IEEEtran}

\IEEEoverridecommandlockouts                              
\usepackage[numbers]{natbib}
\usepackage[bookmarks=true]{hyperref}
\usepackage{graphicx}
\usepackage{amssymb}
\usepackage{amsmath}
\usepackage{amsthm}
\usepackage{amsfonts}
\usepackage{epstopdf}
\usepackage{algorithm}
\usepackage{algorithmic}

\hypersetup{pdfborder={0 0 0},
            breaklinks=true,%
            bookmarks=true,%
            bookmarksnumbered=true,%
            colorlinks=true,%
            citecolor=blue,%
            linkcolor=red,%
}

\newtheorem{theorem}{Theorem}[section]
\newtheorem{remark}{Remark}[section]
\newtheorem{problem}{Problem}[section]
\newtheorem{assumption}{Assumption}[section]
\newtheorem{corollary}{Corollary}[section]
\newtheorem{lemma}{Lemma}[section]

\def\1{\mathbf{1}}

\def\E{\mathbb{E}}
\def\C{\mathbf{C}}

\def\v{{\mathbf v}}

\def\F{{\mathbf F}}
\def\M{{\mathbf M}}

\def\A{{\mathbf A}}
\def\B{{\mathbf B}}
\def\Q{{\mathbf Q}}
\def\R{{\mathbf R}}
\def\H{{\mathbf H}}

\def\Push{{\rm Push}}
\def\Pop{{\rm Pop}}
\def\ProbBound{{\rm PropagateBound}}
\def\Real{{\mathbb R}}
\def\n{{\mathbf n}}

\def\x{{\bf x}}
\def\f{{\mathbf f}}
\def\y{{\mathbf y}}

\def\h{{\mathbf h}}
\def\P{{\mathbf P}}

\def\lmax{\bar{\lambda}}
\def\lmin{\underline{\lambda}}

\newcommand{\map}[3]{{#1}:{#2}\rightarrow{#3}}
\newcommand{\norm}[1]{\|{#1}\|}

\begin{document}

\title{Robust Belief Roadmap:\\ Planning Under Intermittent Sensing}


\author{Shaunak D. Bopardikar \qquad Brendan J. Englot \qquad Alberto Speranzon\thanks{The authors are with United Technologies Research Center. This work was supported by United Technologies Research Center under the Autonomy initiative. Email: \texttt{\{bopardsd, englotbj, sperana\}@utrc.utc.com}.}}


\maketitle

\begin{abstract}
In this paper, we extend the recent body of work on planning under uncertainty to include the fact that sensors may not provide any measurement owing to misdetection. This is caused either by adverse environmental conditions that prevent the sensors from making measurements or by the fundamental limitations of the sensors. Examples include RF-based ranging devices that intermittently do not receive the signal from beacons because of obstacles; the misdetection of features by a camera system in detrimental lighting conditions; a LIDAR sensor that is pointed at a glass-based material such as a window, etc.

The main contribution of this paper is twofold. We first show that it is possible to obtain an analytical bound on the performance of a state estimator under sensor misdetection occurring stochastically over time in the environment. We then show how this bound can be used in a sample-based path planning algorithm to produce a path that trades off accuracy and robustness. Computational results demonstrate the benefit of the approach and comparisons are made with the state of the art in path planning under state uncertainty.
\end{abstract}

\IEEEpeerreviewmaketitle

\section{Introduction}

Map-based, GPS-denied navigation often relies on the measurement of environmental features to perform state estimation.  Whether these features are extracted from camera or LIDAR data, or supplied by range beacons, their measurement will likely be corrupted by noise.  Producing a consistent estimate in the presence of sensor noise is often the primary concern in designing a state estimator, but an important related question is how to navigate robustly when the sensor does not produce a measurement at all, either because the sensor itself is corrupted or environmental conditions prevent the sensor from making a measurement. Examples of such situations include RF-based ranging devices that intermittently do not receive signals from beacons due to obstacles, and misdetection of features by a camera system in textureless areas of the environment or due to adverse lighting conditions.

The main contribution of this paper is twofold. We first show that it is possible to obtain an analytical bound on the performance of a state estimator under probabilistic sensor misdetections. We then show how this bound can be used in a sample-based path planning algorithm that minimizes goal-state uncertainty under such a stochastic mode of misdetection.

Recent work in robotics has emphasized robust path planning under various sources of uncertainty.  The objective is frequently to identify a feasible start-to-goal path that minimizes uncertainty along the path, uncertainty at the goal, or some combination of these and the length of the path.  Actions and measurements affected by noise are often considered, as well as uncertain maps.  The stochastic motion roadmap is a foundational work in modeling path planning under process noise as a Markov decision process (MDP), which is solved optimally using dynamic programming~\cite{alterovitz2007}. Using the framework of Partially Observable Markov Decision Processes (POMDP), Marthi addresses path planning in environments in which obstacles appear/disappear dynamically over time~\cite{marthi2008}.

If stochastic measurements are considered in addition to actions, the planning problem, also a POMDP, is intractable over most state spaces relevant to problems in robotics.  As a result, a variety of algorithms make simplifying assumptions and find high-quality feasible paths that manage uncertainty.  Sample-based motion planning is often utilized to generate a set of collision-free feasible paths from which a minimum-uncertainty path is selected. The belief roadmap (BRM)~\cite{prentice2009} builds a probabilistic roadmap (PRM) \cite{kavraki1996} in a robot's state space, propagates beliefs over the roadmap using an extended Kalman filter (EKF)~\cite{gelb1974}, and plans a path of minimum goal-state uncertainty.  This approach has been extended \cite{he2008} to bias the PRM samples using a Sensory Uncertainty Field (SUF)~\cite{takeda1994}, which expresses the spatial variation in sensor performance over the workspace. Rapidly-exploring random belief trees (RRBTs)~\cite{bry2011} use the EKF to propagate belief states over a rapidly-exploring random graph (RRG)~\cite{karaman2011}, to find asymptotically optimal paths that minimize goal-state uncertainty subject to chance constraints.  Linear quadratic Gaussian motion planning (LQG-MP) \cite{vandenberg2011} pairs an LQG controller-estimator duo with trajectories planned using rapidly-exploring random trees (RRTs)~\cite{lavalle2001}, seeking a path that minimizes the product of collision probabilities at all states.  Instead of sample-based planning, continuous optimization is used by Platt \textit{et al.}~\cite{platt2010}; locally optimal paths are computed directly in belief space under the assumption that the maximum-likelihood measurement is always obtained, and LQG estimation and control are applied.

Some algorithms assume uncertainty in the map used for navigation. Missiuro and Roy~\cite{missiuro2006} employ PRM path planning over a roadmap in which the positions of obstacle vertices are uncertain, and Guibas \textit{et al.}~\cite{guibas2008} extend this approach to 3D workspaces.  This methodology is combined with assumptions of uncertain actions and measurements by Kurniawati \textit{et al.}~\cite{kurniawati2011}, who use a point-based POMDP planner to obtain an approximate minimum-cost solution, where the cost is a combination of movement and collision risk.  A hierarchical approach is adopted by Vitus \textit{et al.}~\cite{vitus2012}, who manage all three sources of uncertainty by decomposing the workspace into a graph and optimizing over the graph in several steps. Wellman et al.~\cite{wellman1995} consider a setting in which the edge costs on the graph are uncertain, with potential probabilistic dependency between the costs, and provide a path planning algorithm that produces optimal paths under time-dependent uncertainty. Acar et al.~\cite{acar2003} present an approach that uses geometric and topological features instead of sensor uncertainty models.

Our problem is related to that of planning over a stochastic map. Uncertainty in the arrangement of obstacles in a map may adversely impact a robot's navigation process, and similarly, the precision of a robot's state estimate will be hindered if its sensors do not detect a measurement. Our aim is to develop a principled method for planning under uncertainty when misdetection by the robot's sensors is a primary concern.
In total, we consider three sources of uncertainty: process noise, sensor noise, and sensor intermittency, for which sensor misdetections occur with a known probability.
Like the BRM method of Prentice and Roy~\cite{prentice2009}, an EKF is used to propagate belief states over a PRM, and a path of minimum goal-state uncertainty is selected.  Performance guarantees are developed that bound the filter's performance under probabilistic misdetections by the sensors, and the likelihood of these misdetections is considered explicitly, along with process and sensor noise, in selecting a minimum-uncertainty path.

Regarding the estimation aspect of the problem, we demonstrate that the choice of the expected maximum eigenvalue of the error covariance matrix as a metric allows us to compute a novel upper bound on its evolution, which is further extended to the case of stochastic misdetections by the robot's sensors. These bounds are distinct from existing results in the literature, surveyed in~\cite{kwon1996}, which are mainly for the Algebraic Riccati equation, representing the steady state value of the expected error covariance instead of its instantaneous value that we are concerned with. Other metrics, such as the trace of the expected error covariance matrix, have been commonly considered in the past, cf.~\cite{prentice2009}. However, to the best of our knowledge, the trace does not offer a tractable means to bound its evolution over time, especially in the stochastic setting of sensor misdetections. Further, while introducing conservativeness in the sense that we consider the maximum mode for the uncertainty, propagating the maximum eigenvalue bound offers a computational advantage since we only need to propagate a scalar quantity instead of the entire covariance matrix.

This paper is organized as follows. In Section~\ref{sec:problem}, we formulate the problem. An analytical bound on the performance of the state estimator with probabilistic sensor misdetections is derived in Section~\ref{sec:analysis}. In Section~\ref{sec:path_planning}, we describe in detail how the analytical bound is used for robust path planning. Computational results are reported in Section~\ref{sec:simulations}. Finally, conclusions and future directions are discussed in Section~\ref{sec:conclusion}.

\section{Problem Formulation}\label{sec:problem}
We consider a general model of an agent whose state evolves as per a non-linear discrete-time dynamical system
\begin{equation}\label{eq:sys}
\x(t+1) = \f(\x(t),\n(t)),
\end{equation}
where $\x \in \Real^{n_x}$ is the state describing the system at time $t$, $\map{\f}{\Real^{n_x}\times\Real^{n_n}}{\Real^{n_x}}$ describes the state transition map of the system and~$\n \in \Real^{n_n}$ is the process noise. The agent is equipped with~$m$ sensors in order to estimate the state~$\x$. Sensors' output is modeled as
\begin{equation}\label{eq:y}
\y_j(t) = \h_j(\x(t), \v_j(t)), \qquad \forall j \in \{1,\dots,m\},
\end{equation}
where $\v_j \in \Real^{n_j}$ is the process noise of the~$j$-th sensor and $\map{\h}{\Real^{n_x}\times\Real^{n_j}}{\Real^{n_{y_j}}}$ describes the relation between state and measurement. We assume that the noise vectors~$\n$ and~$\v_j$ are independently generated mean-zero Gaussian random vectors.

In this paper, we consider situations where sensors can misdetect features and thus, do not produce a measurement at certain time instants. Examples include RF-based signals from beacons that are not detected by the agent because of low SNR, misdetection of features using cameras because of abrupt change of lighting conditions, no LIDAR returns from certain materials, etc.

Under these circumstances, analogous to the Kalman Filter case (see equations 185 and 186 in~\cite{HDW:01}), an Extended Kalman Filter (EKF) based estimator of the state~$\x$ can be written as:
\begin{align}
\mathbf{P}^{-1}_{t+1} &= (\mathbf{F}_{t}\mathbf{P}_{t}\mathbf{F}'_{t} +\Q_{t})^{-1} + \sum_{j=1}^m \gamma_{j,t+1}\H_j' \R_{j,t+1}^{-1} \H_j, \label{eq:KF}\\
\hat {\x}_{t+1} &= \mathbf{P}_{t+1}\Big ( (\mathbf{F}_{t}\mathbf{P}_{t}\mathbf{F}'_{t} + \Q_{t})^{-1} \mathbf{f}(\hat{\x}_{t}, \mathbf{0}) +  \nonumber\\ &\quad
+\sum_{j = 1}^m\gamma_{j,t+1}  \H_j' \mathbf{R}_{j,t+1}^{-1} (\y_{j}(t+1) - \mathbf{h}_j(\mathbf{f}(\hat{\x}_{t}),\mathbf{0})) \Big),\nonumber
\end{align}
where~$\hat{\x}_t$ is the state estimate,~$\P_t$ is the expected error covariance with respect to the process and sensor noise terms, $\mathbf{F}_{t}$ is the linearization of $\mathbf{f}$ around $(\hat{\x}_{t}, \mathbf{0})$ and $\H_j$ is the linearization of $\mathbf{h}_j$ around $(\mathbf{f}(\hat{\x}_{t}), \mathbf{0})$. The matrix~$\Q_{t}$ is the process noise covariance and the~$\R_{j,t+1}$ is the measurement noise covariance associated to the~$j$-th sensor. The variables $\gamma_{j,t+1}$ are binary, $0-1$ stochastic variables that model the misdetection by the~$j$-th sensor.

We are interested to compute a BRM that trades off accuracy and robustness when sensors can stochastically be in misdetection mode during the mission. We will call this \emph{Robust Belief Roadmap} or \emph{RBRM}.

In this work, we make the following assumptions:
\begin{assumption}[Misdetection map]\label{as:knownpdf}
For each sensor $j$, we can characterize the misdetection probability $(1-p_j)$ at each location in the environment.
\end{assumption}
\begin{assumption}[Independence]\label{as:independency}
We assume~$\gamma_{j}$, namely  the misdetections to be independent over time and between sensors. Specifically, the $\gamma_{j}$ are Bernoulli random variables with $p_j$ being the probability of $\gamma_{j} = 1$.
\end{assumption}

For constant values of $p_j$'s, Assumption~\ref{as:independency} is common in the literature pertaining to the research area involving estimation with intermittent observations, e.g., see~\cite{sinopoli2004}. However, most results in this field are concerned with stability of the estimation algorithms, while our focus in this paper is to characterize the evolution of the estimation performance. Further, in our set-up, the parameters $p_j$ of the random variables $\gamma_{j,t}$ in~\eqref{eq:KF} are functions of the state $\x$, whereas in estimation literature, the state dependence is not explicitly considered.

\begin{assumption}[Consistency]\label{as:consistency}
We assume that the state estimate $\hat{\x}_t$ is identical to the nominal trajectory to be followed by the vehicle.
\end{assumption}
The last assumption implies that we have a reasonable nominal model for the motion of the vehicle, and the sensors possess a decent level of accuracy so that the vehicle moves close to the nominal trajectory. This work is concerned with the level of confidence measured through $\P_t$ that we can obtain in our state estimate $\hat{\x}_t$.

With these assumptions, we address the following problems:
\begin{problem}\label{prob:estimation}
Given the sensor accuracies, their misdetection probabilities and a nominal trajectory, determine a bound on the evolution of the expected value of $\mathcal{M}(\P_t)$, where $\mathcal{M}$ represents any function that can capture the uncertainty through $\P_t$. Here, the expectation is taken over the joint distribution of the random variables $\gamma_{j,t}$ representing sensor misdetection.
\end{problem}
Equipped with this theoretical bound, the second goal is to apply the bound to the following problem on path planning under uncertainty.
\begin{problem}
Given a set of candidate trajectories from a start location to a goal location, develop an algorithm that propagates the bound on $\E[\mathcal{M}(\P_t)]$ over the PRM, to output a path having minimum goal-state uncertainty.
\end{problem}

\begin{remark}

\begin{figure}[t!]
\centering
\includegraphics[width=0.95\hsize]{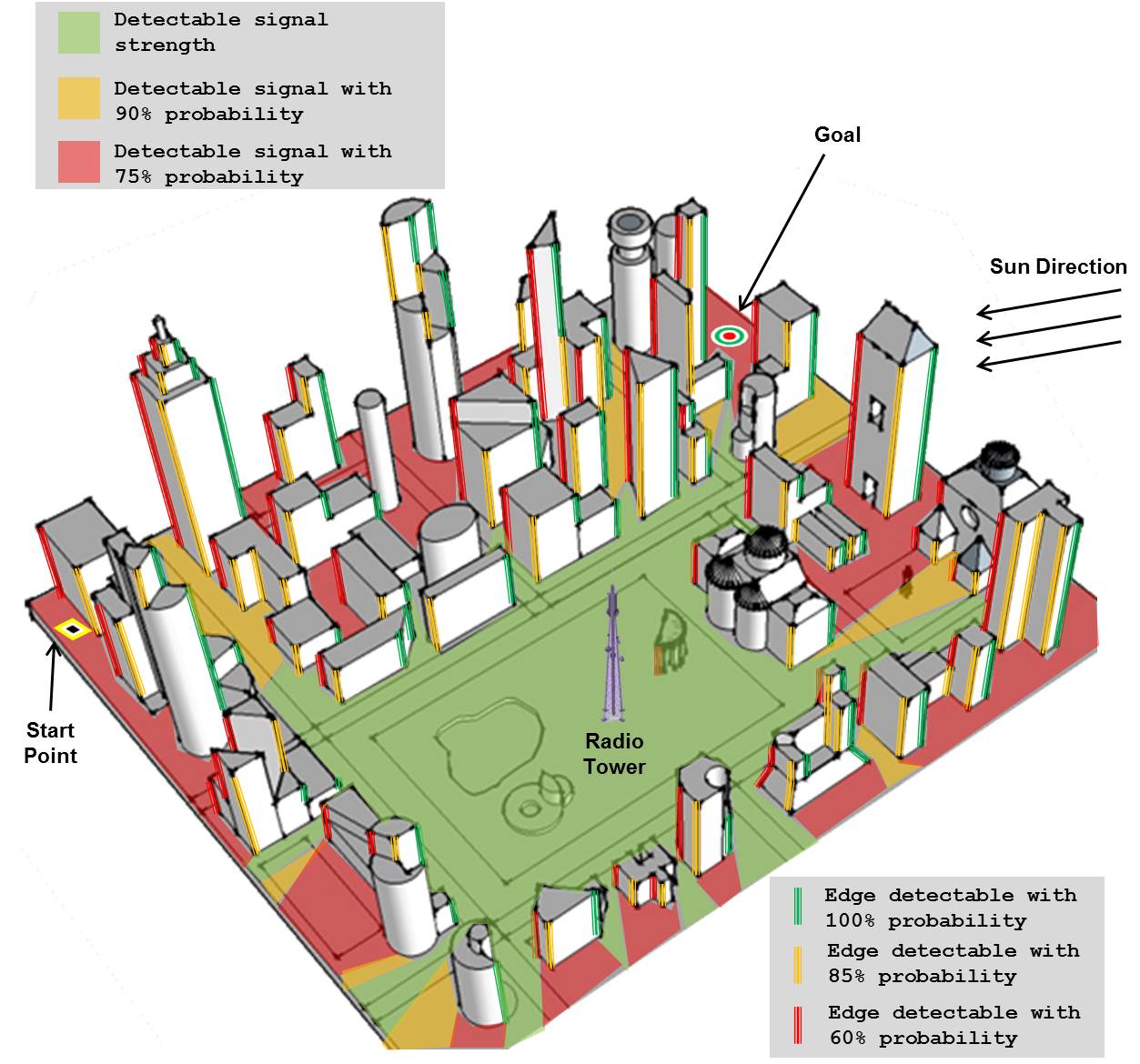}
\caption{Illustration of the model of the environment which could be used for planning. In this case, not only the geometric information about the obstacles and position of beacons is considered, but also the information about the accuracy certain sensors will have in detecting, e.g. edges of buildings and signal from a radio tower (beacon). Note that in this scenario a rather elaborate model is used, where edges are categorized into three classes: detectable, detectable with 85\% and 60\% probability. This different detection accuracy could be caused by the sun's position. Signal strength map from radio tower could be determined using ray tracing algorithms. In this example, just for illustration purposes, we have three regions with different detection probabilities.}\label{fig:Scenario}
\end{figure}

The probability of misdetection of a sensor  (Assumption~\ref{as:knownpdf}) is in general difficult to know precisely. However, thanks to rather realistic simulation environments capable to both simulate sensor responses as well as the environment, one can foresee the possibility of obtaining rather realistic models. For example, given a geometric model of the environment, edges and/or corners of buildings could be marked with different misdetection probabilities if some information from the type of material, the texture, the time of the day the mission is carried out, etc. are considered. Entire areas of the environment could be marked with misdetection probabilities, e.g., modeling the fact that an RF-signal cannot easily be detected behind buildings. This could be obtained with ray tracing based algorithms, see, e.g.,  Figure~\ref{fig:Scenario}.

The misdetection probability of some sensor could also be determined from historical data collected in the mission, for example correlating some information about the environment, such as obstacle density, time of the day the mission was carried out, etc. with the misdetection state of a sensor.

Of course, it is impossible to capture all sources of uncertainty. However, if some of this information is available, the RBRM method can take it into consideration trading off accuracy and robustness.

When such models are only of qualitative nature, the RBRM could be used by the user to assess the robustness of the solution. In practice, the misdetection probabilities can be used as user parameters. Varying such parameters, the user can study how the RBRM changes under, for example, more pessimistic hypothesis on the behavior of certain sensors, having higher misdetection probability in more remote regions of the environment where more uncertainty is expected, etc.
\end{remark}

\section{Analytical Bound on Performance}\label{sec:analysis}

In the seminal work by Prentice and Roy~\cite{prentice2009}, it was shown that if the covariance matrix is factorized as $\P_t = \B_t\C_t^{-1}$, then the time evolution of the terms $\B_t, \C_t$ is linear. This enabled the authors to develop and demonstrate a computationally efficient algorithm to compute a roadmap that captures the estimation accuracy. The function $\mathcal{M}$ used in~\cite{prentice2009} is the trace of the matrix. Leveraging the same factorization, one arrives at the following equation:
\begin{align}\label{eq:roy}
\begin{pmatrix}
\B_t\\ \C_t
\end{pmatrix} =
\begin{pmatrix}
\F_t & \Q_t (\F_t^{-1})' \\%
\M(\gamma_{t})\F_t &
(\F_t^{-1})' + \M(\gamma_{t})\Q_t (\F_t^{-1})'
\end{pmatrix}
\begin{pmatrix}
\B_{t-1} \\ \C_{t-1}
\end{pmatrix},
\end{align}
where~$\M(\gamma_{t}) = \sum_{j=1}^m \gamma_{j,t}\H_j' \R_{j,t}^{-1} \H_j$, which depends on the stochastic variables~$\gamma_{j,t}$. This can be thought as a transfer function that maps the matrices~$\B_t$ and~$\C_t$ from one node of the roadmap to the next~\cite{prentice2009}. In this stochastic setting, however, a direct application of the BRM proposed by Prentice and Roy in~\cite{prentice2009} requires extensive Monte Carlo simulations over all the variables~$\gamma_{j,t}$. Even if the factorization provides a faster computation of the covariance, the method becomes very quickly intractable, especially if~$\gamma_{j,t}$ changes spatially. See, e.g., the signal received from a radio tower as in Figure~\ref{fig:Scenario}.

A way to mitigate this is by taking the expectation with respect to the~$\gamma_{j,t}$, namely by computing~$\E(\B_t)$ and $\E(\C_t)$. This would enable us to compute an expected transfer function between two nodes in the roadmap. However, note that~$\E(\P_t) = \E(\B_t \C_t^{-1}) \neq \E(\B_t)(\E(\C_t))^{-1}$, thus preventing us to compute what the expected state covariance is at each node of the roadmap.

In order to both obtain a meaningful metric on the expected error covariance, which captures the estimate accuracy when there are sensor misdetections, and have computational tractability, we establish a bound on the largest eigenvalue of the covariance. Intuitively, we are approximating the uncertainty at each node of the roadmap with a ball whose radius is the largest eigenvalue of the covariance and we determine a bound on such radius.

\subsection{Bound with all sensors functioning}

In order to derive a bound on the maximum eigenvalue of the expected error covariance, we consider first the case when all the sensors are in working condition at all times. Such analysis is instrumental to derive a bound for the case of sensor misdetection.

Without loss of generality and for the sake of simpler notation, we assume that there exists only one sensor that is in working condition. The bound can be easily generalized to multiple sensors. In the following, we will denote with $\lmin(\A)$ and with $\lmax(\A)$ the minimum and maximum eigenvalue of~$\A$, respectively, where~$\A$ is a positive definite matrix.

For the expected error covariance, the following result provides a recursion to compute an upper bound on $\lmax(\P_t)$.
\begin{theorem}\label{thm:P}
At every time instant $t$,
\begin{equation}\label{eq:lmax}
\lmax(\P_t) \leq \cfrac{\lmax^2(\mathbf{F}_t)\lmax(\mathbf{P}_{t-1})+ \lmax(\mathbf{Q}_t)}{\lmin\left({\H}_t'\mathbf{R}^{-1}_t {\H}_t  \right) (\lmax^2(\mathbf{F}_t)\lmax(\mathbf{P}_{t-1})+ \lmax(\mathbf{Q}_t)) + 1},
\end{equation}
where the Jacobians $\mathbf{F}_t, \mathbf{H}_t$ are evaluated at $\f(\hat{\x}_{t-1},\mathbf{0})$.
\end{theorem}
The proof of Theorem~\ref{thm:P} is reported in the Section~\ref{appx:proofTheorem1} of the Appendix. This result is stated in the form of a recursion mainly because the terms
$\lmax^2(\mathbf{F}_t)$ and ${\H}_t'\mathbf{R}^{-1}_t {\H}_t$ are functions of the estimate $\hat{\x}_{t-1}$. If we can uniformly upper and lower bound them respectively, e.g, in the linear time invariant case, then we can state a uniform upper bound for $\lmax(\P_t)$ as a function of the initial value $\P_0$, given by the next result.

For this result, we introduce the following notation. Let $\mathcal{X}:= \{\hat{\x}_0, \hat{\x}_1, \dots, \hat{\x}_T\}$, denote a set of estimates at different times. Under Assumption~\ref{as:consistency}, this set is identical to the nominal trajectory sampled at the corresponding times. Define,
$$
\mathcal{X}_S := \Big \{ \hat{\x} \in \mathcal{X} \, : \lmin\Big(\H(\f(\hat{\x},\mathbf{0}))' \mathbf{R}_t^{-1} \H(\f(\hat{\x},\mathbf{0})) \Big) = 0 \Big\},
$$
denote the subset of $\mathcal{X}$ at which the sensor provides no useful information. The filter runs open-loop at these locations.

Then, the following result holds.

\begin{theorem}\label{thm:apriori}
Suppose that the cardinality of $\mathcal{X}_S$ is $\kappa$.
 Then, under Assumption~\ref{as:consistency}, at the final estimation time instant $T$,
\begin{multline*}
\lmax(\mathbf{P}_{T}) \leq b\sum_{j=1}^{\kappa}a^{j-1} - a^{\kappa}\zeta + a^{\kappa}\Big / \\
\left ( \left( \frac{d-\zeta  c}{\zeta c +a} \right)^{T-\kappa} \frac{1}{\zeta + \lmax(\mathbf{P}_0)} + \frac{c}{\zeta c +a} \left(\frac{1-\frac{( d -\zeta c)^{T-\kappa}}{(\zeta c +a)^{T-\kappa}}}{1-\frac{(d-\zeta c)}{(\zeta c +a)}} \right) \right ),
\end{multline*}
where
\begin{align*}
a &:= \sup_{\hat{\x} \in \mathcal{X}}\lmax(\mathbf{F}(\hat{\x}))\,, \qquad b := \sup_t \lmax(\mathbf{Q}_t)\,, \\
c&:=\inf_{\hat{\x}\in \mathcal{X}\setminus \mathcal{X}_S}\lmin\left(\H_{j}(\f(\hat{\x},\mathbf{0}))' \mathbf{R}_{j,t}^{-1} \H_{j}(\f(\hat{\x},\mathbf{0})) \right)\,, \\
d &:=  b  c/a + 1\,, \qquad
\zeta := ( d - a + \sqrt{( d - a)^2 + 4 b c})/(2 c)\,.
\end{align*}
\end{theorem}
The proof of this result is presented in Section~\ref{appx:proofTheorem_aprori} of the Appendix. To extend both of these results for $m$ functioning sensors, we simply replace the term ${\H}_t'\mathbf{R}^{-1}_t {\H}_t$ by $\sum_{j=1}^m{\H}_{j,t}'\mathbf{R}_{j,t}^{-1} {\H}_{j,t}$.

\subsection{Bound under Stochastic Sensor Misdetections}
In this section, we extend the analysis to the case of multiple sensors, added to improve robustness, but that can produce misdetections as per Assumptions~\ref{as:knownpdf} and~\ref{as:independency}. The metric which we analyze is $\E[\lmax(\P_t)]$, where the expectation is taken over the stochastic process of sensor misdetections.

For brevity, let $\ell_t := \lmax(\P_t)$. For $k \in \{1,\dots, m\}$, define:
\begin{align}\label{eq:cd}
c_{i_1, \dots, i_k} &= a\lmin\Big(\sum_{j = 1}^{k} \H_{i_j,t}'\R_{i_j,t}^{-1} \H_{i_j,t}\Big), \nonumber \\
d_{i_1, \dots, i_k} &= bc_{i_1, \dots, i_k}/a + 1,
\end{align}
where a tuple $i_1,\dots, i_k$ is a subset of $\{1,\dots, m\}$, and $a, b$ are as defined in Theorem~\ref{thm:apriori}.

Using this notation, we have the following recursion which provides an upper bound on $\E[\ell_t]$, referred to hereafter as $\overline{\E[\ell_t]}$. 

\begin{theorem}[Stochastic misdetections]\label{thm:stochastic}
Under Assumptions~\ref{as:knownpdf} and~\ref{as:independency}, at any given time instant~$t$,  $\overline{\E[\ell_t]}$ generated as per the following recursion,
\begin{align*}
&\overline{\E[\ell_{t}]} = (a \overline{\E[\ell_{t-1}]} + b) \Big( (1-p_1)\dots(1-p_m) \\
&+\frac{p_1(1-p_2)\dots(1-p_m)}{c_1\overline{\E[\ell_{t-1}]} + d_1} + \dots + \frac{(1-p_1)(1-p_2)\dots p_m}{c_m\overline{\E[\ell_{t-1}]} + d_m} \\
&+ \frac{p_1p_2\dots(1-p_m)}{c_{12}\overline{\E[\ell_{t-1}]} + d_{12}} + \dots + \frac{(1-p_1)\dots p_{m-1}p_m}{c_{m-1,m}\overline{\E[\ell_{t-1}]} + d_{m-1,m}} \\
&\:\:\vdots \\
&+ \frac{p_1\dots p_m}{c_{1,\dots, m} \overline{\E[\ell_{t-1}]} + d_{1,\dots, m}} \Big).
\end{align*}
is an upper bound on $\E[\ell_t]$.
\end{theorem}

The proof of this result is presented in Section~\ref{appx:proofTheorem2} of the Appendix.

This bound requires the enumeration of all of the~$2^m$ possibilities of sensor combinations, and therefore, the computational complexity scales undesirably with~$m$. One way to derive an efficient bound is to obtain a uniform lower bound $\bar c$ on each of the c's. In that case, the common denominator of the right hand side terms becomes $\bar c\:\E[\ell_{t-1}] + \bar d$. The recursion then simplifies to
\[
\overline{\E[\ell_t]} = (a\overline{\E[\ell_{t-1}]} + b) \left(\prod_{j=1}^m(1-p_j) + \frac{1 - \prod_{j=1}^m(1-p_j)}{\bar c\,\overline{\E[\ell_{t-1}]} + \bar d} \right ).
\]
This recursion can be evaluated on similar lines to the proof of Theorem~\ref{thm:apriori} to obtain a $\overline{\E[\ell_t]}$ as a function of $\E[\ell_0]$.

For certain types of sensor suites, one may be able to derive a slightly conservative, but computationally efficient upper bound which we report next.

\begin{corollary}[Simplified bound]\label{thm:conservative_bound}
Under  Assumptions~\ref{as:knownpdf} and~\ref{as:independency}, at any given time instant~$t$, $\overline{\E[\ell_t]}$ generated as per the following recursion,
\begin{multline*}
\overline{\E[\ell_{t}]} = (a\overline{\E[\ell_{t-1}]} + b) \left(\prod_{j=1}^m(1-p_j)
+ \sum_{j = 1}^m \frac{p_j}{c_j\overline{\E[\ell_{t-1}]} + d_j} \right).
\end{multline*}
is an upper bound on $\E[\ell_t]$.
\end{corollary}
The proof is reported in the Section~\ref{appx:proofTheorem3} of the Appendix. The main advantage of this bound is the computational efficiency as compared to the one in Theorem~\ref{thm:stochastic}, since this needs to evaluate only $m$ terms. However, this bound requires at least one of the sensors to have its $c$ value to be strictly positive, and therefore, may become too conservative. We will use Theorem~\ref{thm:stochastic} in our proposed RBRM approach.

\section{Application to Path Planning Missions}\label{sec:path_planning}

The upper bound on $\E[\ell_t]$ given in Theorem \ref{thm:stochastic} may be used to plan paths of minimum expected goal-state uncertainty in a manner similar to the belief roadmap algorithm \cite{prentice2009}.  We will assume that a probabilistic roadmap with node set $N$ and edge set $E$ is provided as input, along with beliefs $\mu_0$ and $\mu_{goal}$ defining a start state and goal state on the roadmap. We also assume that for every node $n \in N$, the triple $n = \{\mu,\overline{\E[\ell]}, \pi\}$ is stored, which contains the belief, the eigenvalue bound, and the path $\pi$ (beginning at $\mu_0$) associated with this node. We refer to individual members of the triple using the notation $n[\mu]$, $n[\overline{\E[\ell]}]$, and $n[\pi]$.  Belief propagation and graph search proceeds similarly to that of the standard BRM algorithm; $\overline{\E[\ell_t]}$ is propagated according to the recursive inequality given in Theorem \ref{thm:stochastic}, and is used in place of the nominal-trajectory expected error covariance matrix that is propagated in the standard BRM.  We assume the bound is used to compute a transfer function $\overline{\E[\ell]}_l = \zeta(i,l,\overline{\E[\ell]}_i)$, that takes as input the indices of an edge $e_{i,l}$ in the roadmap, and the eigenvalue bound associated with node $n_i$.  In the context of the graph search, we treat $\overline{\E[\ell]}$ independently of time, and assume that $n_i[\overline{\E[\ell]}]$ represents the best-yet covariance eigenvalue bound identified at $n_i$. The search process is shown in Algorithm \ref{alg1}.

\begin{algorithm}
\caption{$ n_{goal}\left[\pi\right] = RBRM(\mu_0,\mu_{goal},\overline{\E[\ell]}_{0},N,E)$}
\label{alg1}
\begin{algorithmic}
\FOR{ $ e_{i,l} \in E $ }
\STATE $\xi(i,l,\overline{\E[\ell]}_i) \leftarrow \ProbBound(e_{i,l})$
\ENDFOR
\STATE $Q \leftarrow n_0 = \{ \mu_0,\overline{\E[\ell]}_0,\emptyset \}$
\WHILE{ $Q \neq \emptyset$}
\STATE $ n_i \leftarrow \Pop(Q)$
\FOR{$ n_l \in e_{i,l}$}
\IF{$n_l \notin n_i\left[\pi\right]$}
\STATE $\overline{\E[\ell]}_l \leftarrow \xi(i,l,n_i[\overline{\E[\ell]}])$
\IF{$\overline{\E[\ell]}_l < n_l[\overline{\E[\ell]}]$}
\STATE $n_l \leftarrow \{ n_l[\mu], \overline{\E[\ell]}_l, n_i[\pi] \cup n_l \}$
\STATE $Q \leftarrow \Push(Q,n_l)$
\ENDIF
\ENDIF
\ENDFOR
\ENDWHILE
\RETURN$ n_{goal}[\pi]$
\end{algorithmic}
\end{algorithm}

The use of our proposed approach, i.e., propagation of $\overline{\E[\ell_t]}$, provides us with significant computational advantage over existing methods such as~\cite{prentice2009}. If we were to use their factorization from~\eqref{eq:roy}, then we would have to compute: 1) $2^m$ realizations of the matrix $\B_t$ (one for each subset of misdetecting sensors), 2) inverses of $2^m$ realizations of the matrix $\C_t$, 3) multiply each realization of $\B_t$ with corresponding $\C_t^{-1}$ and finally, 4) sum up the $2^m$ terms to compute $\E[\mathbf{P}_t]$ or its trace. Instead, our approach requires the computation of minimum eigenvalues of $2^m$ much smaller sized matrices, i.e., sums of the terms $\H_j'\mathbf{R}_{j,t}^{-1} \H_j$, for which efficient algorithms exist even for larger sizes~\cite{okamoto1993}, along with step 4) of the above, which provides significant savings in high dimensional state space.

\section{Computational Results}\label{sec:simulations}

\begin{figure}[t]
\centering
\includegraphics[width=.96\columnwidth]{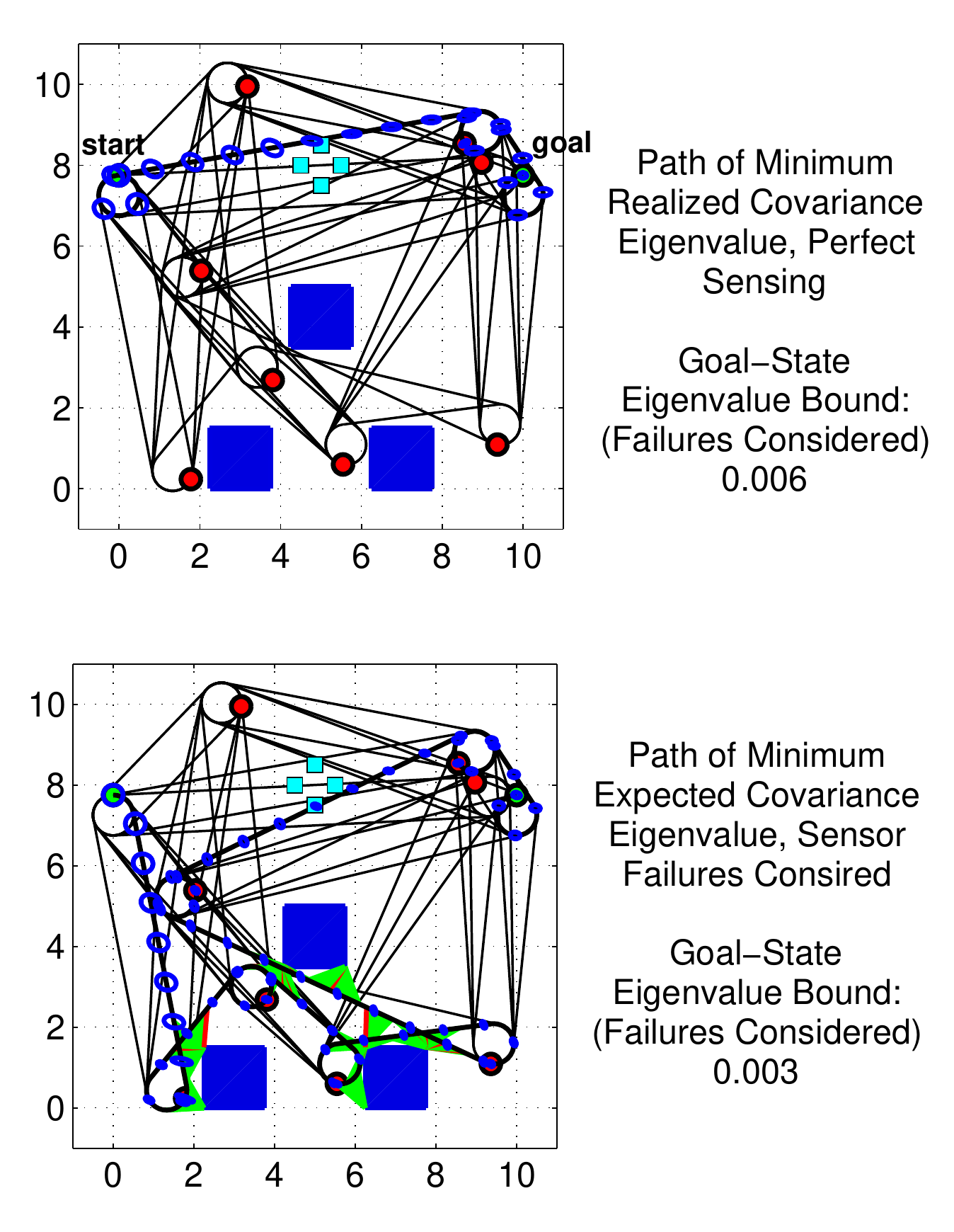}
\caption{Planned paths in a workspace populated with obstacles (measured by laser) and UWB beacons. The robot receives the beacon measurements with probability 0.1, and extracts obstacle corners from laser data with probability 0.9. At top, a path planned using  $\ell_t$ as a performance metric, neglecting all probabilistic sensor misdetections.  At bottom, a path planned using $\overline{\E[\ell_t]}$ as a performance metric, which considers the misdetection probability of each sensor.  The UWB beacons are queried at every measurement iteration; the laser has a range of one unit and its planned measurements are rendered (green for a successful measurement and red for a misdetection) for a representative failure scenario. Ninety-five percent confidence covariance ellipses are plotted at regular intervals along each path.}
\label{fig:planned_paths_two_sensors}
\end{figure}

\begin{figure}[t]
\centering
\includegraphics[width=.96\columnwidth]{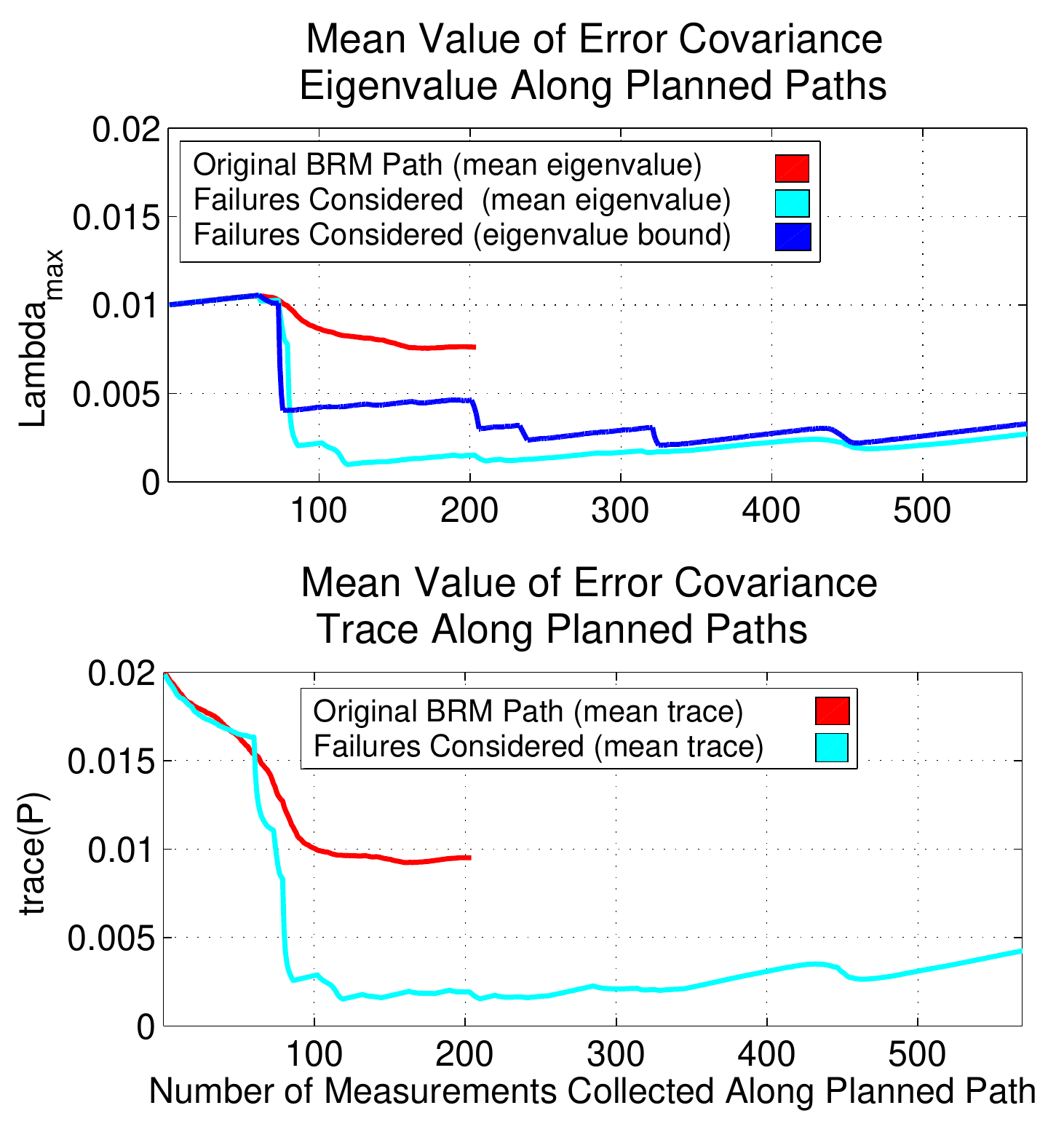}
\caption{At top, the propagation in time of the eigenvalue performance metrics over the paths in Figure \ref{fig:planned_paths_two_sensors}. At bottom, $tr(\P_t)$ is also given for both paths.  All quantities except $\overline{\E[\ell_t]}$ represent the mean over one hundred Monte Carlo trials in which different sequences of sensor misdetections occur according to the prescribed probabilities.}
\label{fig:metrics_two_sensors}
\end{figure}

We next plan paths of minimum uncertainty under process noise, sensor noise, and probabilistic misdetections for a planar Dubins vehicle \cite{dubins1957} in an environment populated with obstacles.  We assume a robot is using three sensors for navigation: ultra-wideband (UWB) range beacons, a laser rangefinder for measuring obstacle vertices, and odometry that is subject to drift over time.  The beacons provide measurements throughout the workspace, but their noise properties are assumed to vary as a function of distance to the robot\footnote{A more general model could also consider a bias term as described in~\cite{prentice2009}. This could be easily added also in our framework.}, according to
\begin{align}\label{eq:v}
\v_j(t) \sim \mathcal{N}(0,\sigma(d_j(t))^2)\,, \\
\sigma(d_j(t)) = \alpha d_j(t) + \sigma_0\,.
\end{align}
The noise associated with the range measurement of beacon~$j$ has a standard deviation that varies linearly in the Euclidean distance $d_j(t)$ between the robot and sensor~$j$.  The standard deviation takes on value $\sigma_0$ at range zero and increases according to the coefficient~$\alpha$.  For the laser rangefinder, we assume the measurement of range to an obstacle vertex is corrupted by Gaussian white noise with properties that do not vary spatially, and the vertices measured are always correctly associated with a prior map.  The maximum range of the laser is limited, however, and obstacles can only be detected in close proximity to the robot.

A start state and goal state are designated for the robot, and a PRM is used to identify feasible paths between the start and goal.  To select the path of minimum goal-state uncertainty, two methodologies are compared: the original BRM algorithm, with no notion of sensor misdetections, and the proposed RBRM algorithm, which uses $\overline{\E[\ell_t]}$ as a cost metric instead of $tr(\P_t)$.  
For all path planning scenarios investigated, the standard BRM algorithm was found to choose the same path regardless of whether $tr(\P_t)$ is used as the cost metric or $\ell_t$ is used instead.  Evaluating $\ell_t$ over the roadmap offers a better comparision with $\overline{\E[\ell_t]}$, and so both $tr(\P_t)$ and  $\ell_t$ are computed for comparison with $\overline{\E[\ell_t]}$ in the results to follow.

The first scenario considered is illustrated in Figure \ref{fig:planned_paths_two_sensors}, in which a robot must plan from start to goal in a workspace populated with three obstacles and four range beacons.  Very simple collision-free paths are evident through the upper reaches of the workspace, but in cases where the probability of UWB misdetection is high, it is advantageous for the robot to travel through the obstacles to collect laser measurements that reduce position uncertainty.  For the specific case plotted in Figure \ref{fig:planned_paths_two_sensors}, the beacons have a ten percent probability of delivering a successful measurement to the robot, and the laser (with a maximum range of one unit) has a ninety percent probability of successfully extracting an obstacle vertex and measuring its range to the robot.  When intermittent sensing is neglected, the robot takes a short path to the goal that collects measurements from the UWB beacons only.  When intermittent sensing is considered, the robot takes a detour through the obstacles to reduce the uncertainty of its state estimate.  For both paths, one hundred Monte Carlo simulations were performed in which sensors fail according to the prescribed probabilities, and the resulting mean values of $tr(\P_t)$ and $\ell_t$ are compared with $\overline{\E[\ell_t]}$ in Figure \ref{fig:metrics_two_sensors}.

\begin{figure}[t]
\centering
\includegraphics[width=.96\columnwidth]{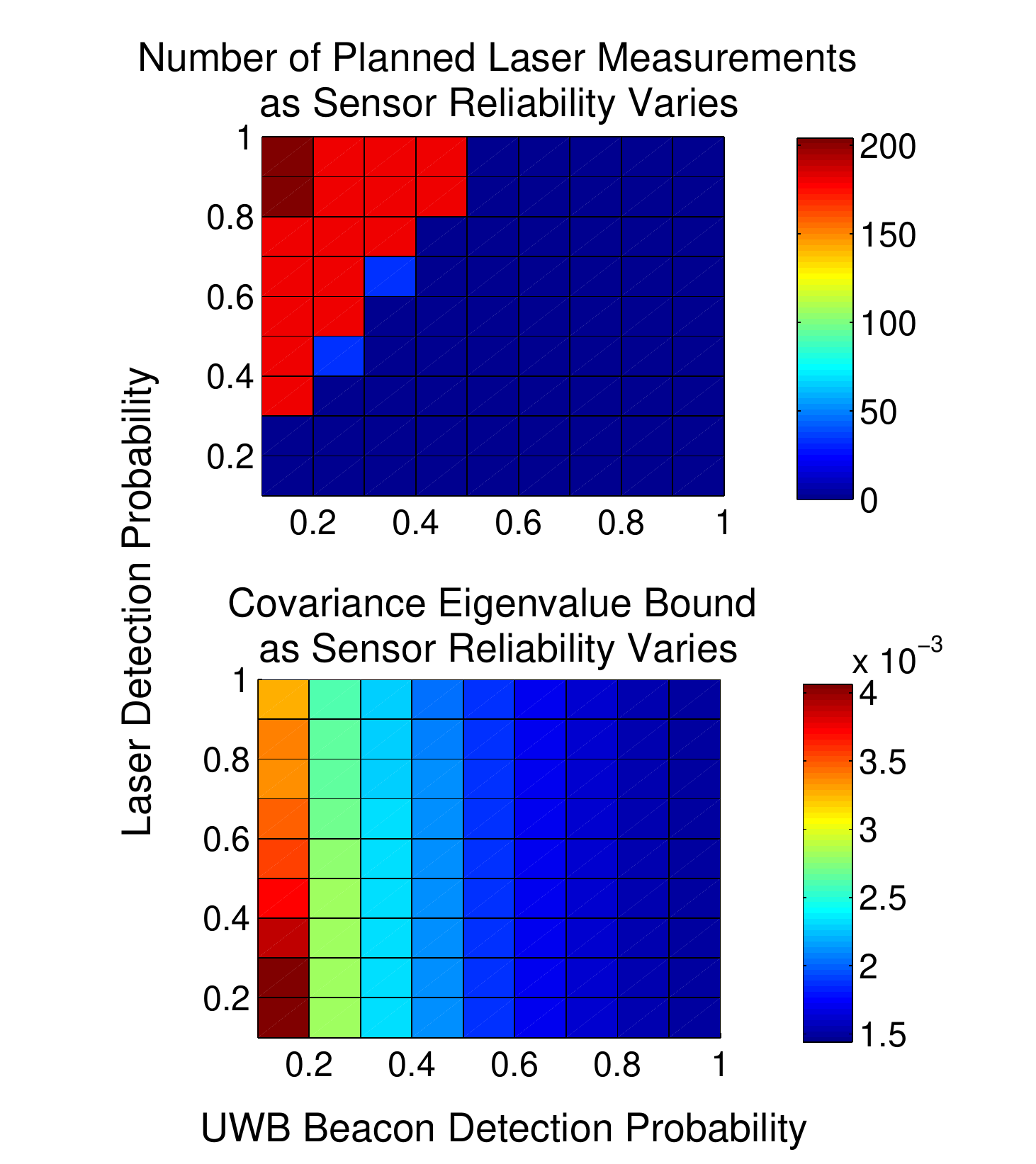}
\caption{Characteristics of a planned path are plotted as a function of sensor reliability, for the two-sensor example shown in Figure \ref{fig:planned_paths_two_sensors}.  At top, the covariance eigenvalue bound at the goal state is illustrated, and at bottom, the number of planned laser measurements along the selected path is illustrated.  The upper left corner of each plot corresponds to the parametrization used in Figures \ref{fig:planned_paths_two_sensors} and \ref{fig:metrics_two_sensors}. }
\label{fig:lambda_grids}
\end{figure}

This planning scenario is next considered over a range of different misdetection probabilities, for both the laser and the UWB beacons, and the results are summarized in Figure \ref{fig:lambda_grids}. The number of planned laser measurements in the minimum uncertainty path, computed using $\overline{\E[\ell_t]}$, is given at top, and the value of $\overline{\E[\ell_t]}$ at each path's goal state is given at bottom.  The zero-range noise level $\sigma_0^2$ selected for the UWB beacons is an order of magnitude lower than the constant variance representing the laser noise, and so the UWB beacons are used exclusively for all scenarios in which they are more than fifty percent reliable, even if the laser is more reliable.

\begin{figure}[t]
\centering
\includegraphics[width=.96\columnwidth]{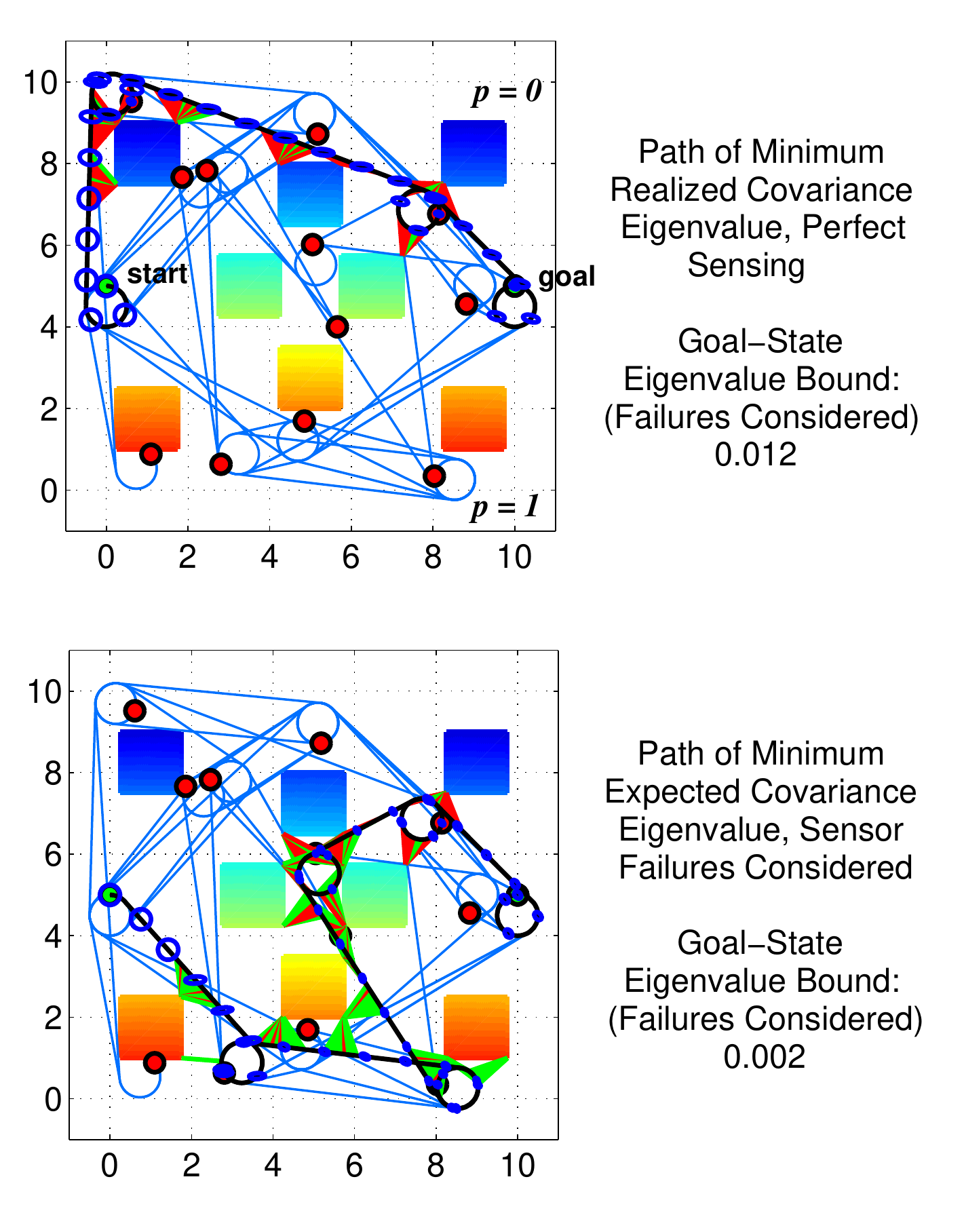}
\caption{Planned paths in a workspace over which the probability of a successful corner detection varies spatially along the vertical axis. Obstacles measured at bottom have the highest probability of a successful measurement, and obstacles measured at top have a near-zero probability of a successful measurement. At top, a path planned using $\ell_t$ as a performance metric, neglecting all probabilistic sensor misdetections.  At bottom, a path planned using $\overline{\E[\ell_t]}$ as a performance metric, which considers the misdetection probability of each sensor. The laser has a range of one unit and its planned measurements are rendered (green for a successful measurement and red for a misdetection) for a representative scenario. Ninety-five percent confidence covariance ellipses are plotted at regular intervals along each path.}
\label{fig:planned_paths_laser_only}
\end{figure}

\begin{figure}[t]
\centering
\includegraphics[width=.96\columnwidth]{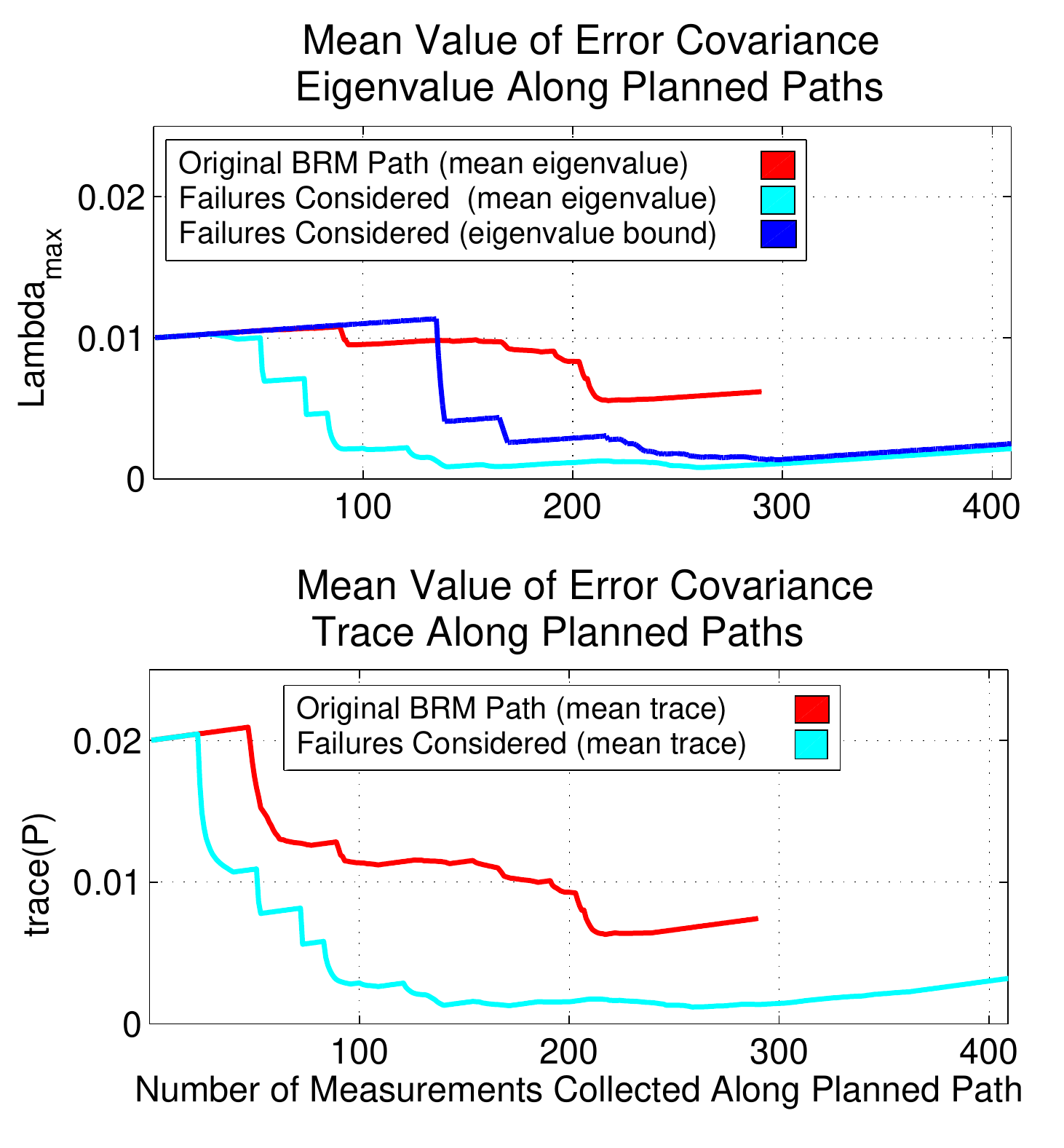}
\caption{At top, the propagation in time of the eigenvalue performance metrics over the paths in Figure \ref{fig:planned_paths_laser_only}.  At bottom, the trace of the expected error covariance matrix is given for both paths.  All quantities except the eigenvalue bound represent the mean over one hundred Monte Carlo trials in which different sequences of sensor misdetections occur according to the prescribed probabilities.}
\label{fig:metrics_laser_only}
\end{figure}

A final path planning test case with continuously varying sensor intermittency is considered in Figure \ref{fig:planned_paths_laser_only}.  In a workspace populated with eight obstacles and no UWB beacons, we assume that a light source causes the expected sensing intermittency to vary continuously along the vertical axis of the workspace, with a high detection probability at bottom and a low detection probability at top.  Neglecting sensor intermittency, the standard BRM algorithm plans a path through the upper region of the workspace, and considering sensor intermittency, planning with $\overline{\E[\ell_t]}$ yields a path that collects many high-probability measurements from the lower region of the workspace to minimize uncertainty at the goal state.  The candidate metrics, averaged over one hundred simulated cases of sensor intermittency, are given in Figure \ref{fig:metrics_laser_only}.

\section{Conclusion and Future Directions} \label{sec:conclusion}

In this paper we have described how to plan when sensors used for state estimation are not only noisy but may fail to produce measurements because of misdetections. Being able to tradeoff both accuracy and robustness is very appealing as autonomous vehicles heavily rely on complex sensors such as cameras and LIDAR whose capability of extracting relevant information, such as features and point clouds, strongly depends on environmental information that can be predicted to a certain extent, such effect of lighting conditions, type of surfaces, etc. Even in the case when such information is not fully available, the proposed methodology can be very beneficial to study the robustness of the path to intermittent sensing, by testing the robust roadmap, for example, by choosing various probabilities of intermittency at various location in the map.

Future directions include the incorporation of map uncertainty within the current framework and the use of the current analysis for multi-objective path planning.

\bibliographystyle{plainnat}
\bibliography{references}

\appendix
In this appendix, we have included the proofs of all technical results presented in this paper.

\section{Proof of Theorem~\ref{thm:P}}
\label{appx:proofTheorem1}

To prove Theorem~\ref{thm:P}, we will use the following set of matrix identities. Although these inequalities are fairly standard, we provide a proof for the sake of completeness.
\begin{lemma}[\cite{Horn1985}] \label{lemma:eigs} Given $n \times n$ positive definite matrices $X, Y, Z$,
\begin{align}
\label{eq:inv}
\lmin(X+Y) &\geq \lmin(X) + \lmin(Y) \\ \label{eq:invtrace}
\lmax(X+Y) &\leq \lmax(X) + \lmax(Y)  \\ \label{eq:trace_of_prod}
\lmax(XY) &\leq \lmax(X)\lmax(Y)
\end{align}
where $\lmin,\lmax$ denote the minimum and the maximum eigenvalues of the matrix, respectively.
\end{lemma}
\begin{IEEEproof}
We prove only the max. The min part follows analogously by reversing signs. To prove \eqref{eq:invtrace}, recall that
\begin{align*}
\lmax(X+Y) &= \max_{x \neq 0} \frac{\norm{(X+Y)x}}{\norm{x}}
\leq  \max_{x \neq 0} \frac{\norm{Xx}}{\norm{x}} +\max_{x \neq 0} \frac{\norm{Yx}}{\norm{x}}  \\
&= \lmax(X) + \lmax(Y),
\end{align*}
where we used the triangle inequality in the second step. \eqref{eq:trace_of_prod} follows by using the same steps, except that we use the sub-multiplicativity of the norm to obtain the inequality.
\end{IEEEproof}

We are now ready to establish Theorem~\ref{thm:P}.

\begin{IEEEproof}[Proof of Theorem~\ref{thm:P}] For the recursion, we use the following \emph{inverse covariance form}~\cite{HDW:01}
$$
\mathbf{P}_t^{-1} = (\mathbf{F}_{t-1}\mathbf{P}_{t-1}\mathbf{F}_{t-1}' + \mathbf{Q}_{t-1})^{-1} +  \H_t'\mathbf{R}_{t}^{-1} \H_t,
$$
where $\mathbf{F}, \H$ are Jacobians of $\mathbf{f}, \mathbf{h}$ around the state prediction at time $t-1$. Using the bounds in Lemma~\ref{lemma:eigs}, we obtain
\begin{align*}
&\lmax(\mathbf{P}_t) = \lmax\Big (\Big [{\H}_t'\mathbf{R}_{t}^{-1} {\H}_t   + (\mathbf{F}_{t-1}\mathbf{P}_{t-1}\mathbf{F}_{t-1}' + \mathbf{Q}_{t-1})^{-1} \Big ]^{-1} \Big ) \\
&= \cfrac{1}{\lmin\Big ( {\H}_t'\mathbf{R}_{t}^{-1} {\H}_t   + (\mathbf{F}_{t-1}\mathbf{P}_{t-1}\mathbf{F}_{t-1}' + \mathbf{Q}_{t-1})^{-1} \Big ) }\\
&\overset{\eqref{eq:inv}}\leq \cfrac{1}{\lmin\left( {\H}_t'\mathbf{R}_{t}^{-1} {\H}_t   \right) + \lmin((\mathbf{F}_{t-1}\mathbf{P}_{t-1}\mathbf{F}_{t-1}' + \mathbf{Q}_{t-1})^{-1} ) }\\
&= \cfrac{1}{\lmin\left({\H}_t'\mathbf{R}_{t}^{-1} {\H}_t  \right)+ (\lmax(\mathbf{F}_{t-1}\mathbf{P}_{t-1}\mathbf{F}_{t-1}' + \mathbf{Q}_{t-1}))^{-1} } \\ &\overset{\eqref{eq:invtrace},\eqref{eq:trace_of_prod}}\leq
\cfrac{\lmax^2(\mathbf{F}_{t-1})\lmax(\mathbf{P}_{t-1})+ \lmax(\mathbf{Q}_{t-1})}{\lmin\left({\H}_t'\mathbf{R}_{t}^{-1} {\H}_t  \right) (\lmax^2(\mathbf{F}_{t-1})\lmax(\mathbf{P}_{t-1})+ \lmax(\mathbf{Q}_{t-1})) + 1}.
\end{align*}
\end{IEEEproof}

\section{Proof of Theorem~\ref{thm:apriori}}
\label{appx:proofTheorem_aprori}

To prove Theorem~\ref{thm:apriori}, we require the following intermediate result, which establishes a bound for the value of a scalar variable $l$ which may evolve as per one out of two equations at any given time.

\begin{lemma}\label{lem:kappa}
Suppose that in the time interval $[0, 1, \dots, T]$, a scalar variable $\ell$ evolves as per
\[
\ell_{t+1} = \begin{cases} \cfrac{a \ell_t +  b}{c \ell_t +  d}, &\text{for some } T-\kappa \text{ instants,}\\
al_t + b, &\text{for the remaining } \kappa \text{ instants,}\end{cases}
\]
where $a, b, c, d$ are some finite positive scalars, then
\begin{multline*}
\ell_T \leq b\sum_{j=1}^{\kappa}a^{j-1} - \zeta a^{\kappa} + a^{\kappa}\Big / \\
\left ( \left( \frac{d-\zeta c}{\zeta c +a} \right)^{T-\kappa} \frac{1}{\zeta + l_0} + \frac{ c}{\zeta c +a} \left( \frac{1-\frac{(d -\zeta c)^{T-\kappa}}{(\zeta  c +a)^{T-\kappa}}}{1-\frac{( d-\zeta c)}{(\zeta c +a)}} \right) \right )
\end{multline*}
where $\zeta$ is defined in Theorem~\ref{thm:P}.
\end{lemma}

\begin{IEEEproof}
Observing that for the above evolution of $l$,
\[
\frac{al + b}{c l + d} + b \geq \frac{a(al +  b) +  b}{c l + d},
\]
which means that for \emph{any} sequence of $\kappa$ occurrences of the second equation, one can always upper bound the resulting $l$ trajectory by considering all the occurrences of evolution by the first equation, followed by the second.

The evolution given by the first equation in the time interval $[0, T-\kappa]$ can be simplified as follows.  Set
\begin{align*}
  \mu_t &:= \frac{1}{\zeta + \ell_t}
\end{align*}
Since $b c > 0$,  we get
\begin{align*}
\mu_t &\geq \frac{d-\zeta  c}{\zeta c + a} \mu_{t-1} + \frac{ c}{\zeta c + a} \\
\Rightarrow \mu_t &\geq \Big ( \frac{ d-\zeta  c}{\zeta c + a} \Big )^{k} \mu_0  \\
&+ \frac{ c}{\zeta c +a} \Big( \frac{1-(d -\zeta c)^{k}/(\zeta c + a)^{k}}{1-(d-\zeta  c)/(\zeta c +a)} \Big),
\end{align*}

Therefore,
\begin{multline*}
\ell_{T-\kappa} \leq 1 \Big / \\
\left ( \left( \frac{d-\zeta c}{\zeta c +a} \right)^{T-\kappa} \frac{1}{\zeta + l_0} + \frac{ c}{\zeta c +a} \left( \frac{1-\frac{( d -\zeta c)^{T-\kappa}}{(\zeta  c +a)^{T-\kappa}}}{1-\frac{( d-\zeta c)}{(\zeta c +a)}} \right) \right ) - \zeta.
\end{multline*}
The claim now follows since $l_T$ can be at most the above right hand side subject to the second, linear evolution for $\kappa$ time steps.
\end{IEEEproof}

We can now prove Theorem~\ref{thm:apriori}.

\begin{IEEEproof}[Proof of Theorem~\ref{thm:apriori}]
Consider the recursion from Theorem~\ref{thm:P}. Substituting $z_t := \lmax(\mathbf{P}_t)$, we obtain the following linear rational recurrence,
\begin{equation}\label{eq:riccatti}
z_t \leq \cfrac{az_{t-1} +  b}{ c z_{t-1} + d}
\end{equation}
where we used the definition of $ a, b,  c$ and $ d$.

Now, whenever $\hat x \in \mathcal{X}\setminus \mathcal{X}_S$, $z$ will evolve as per~\eqref{eq:riccatti}. Otherwise, $z$ evolves as per
\[
z_t \leq a z_{t-1} +  b,
\]
which happens at most $\kappa$ times as per the assumption. Therefore, applying Lemma~\ref{lem:kappa}, the claim is established.
\end{IEEEproof}

\section{Proof of Theorem~\ref{thm:stochastic}}
\label{appx:proofTheorem2}

\begin{IEEEproof}[Proof of Theorem~\ref{thm:stochastic}]
Conditioning on the value of $\ell_{t-1}$, we can write the following equality upon enumerating all possibilities of the sensors misdetecting.
\begin{align*}
&\E[\ell_{t} | \ell_{t-1}] = (a \ell_{t-1} + b) \Big( (1-p_1)\dots(1-p_m) \\
&+\frac{p_1(1-p_2)\dots(1-p_m)}{c_1\ell_{t-1} + d_1} + \dots + \frac{(1-p_1)(1-p_2)\dots p_m}{c_m\ell_{t-1} + d_m} \\
&+ \frac{p_1p_2\dots(1-p_m)}{c_{12}\ell_{t-1} + d_{12}} + \dots + \frac{(1-p_1)\dots p_{m-1}p_m}{c_{m-1,m}\ell_{t-1} + d_{m-1,m}} \\
&\vdots \\
&+ \frac{p_1\dots p_m}{c_{1,\dots, m} \ell_{t-1} + d_{1,\dots, m}} \Big).
\end{align*}
Now, the function $g(x) := (ax + b)/(cx+d)$, is concave in $x \in \Real^+$ for any positive $c,d$ such that $ad > bc$ (This can be checked by computing $\partial^2 g/\partial x^2$). Therefore, from~\eqref{eq:cd}, each of the summing terms on the right hand side of the above equality are concave in the argument $\ell_{t-1}$. Unconditioning on the random variable $\ell_{t-1}$, and applying Jensen's inequality\footnote{Jensen's inequality: For a random variable $X$ with finite expectation, and a concave, real-valued function $\phi(X)$, $\E[\phi(X)] \leq \phi(\E[X])$.}, we obtain 
\begin{align}\label{eq:ell_ineq}
&\E[\ell_{t}] \leq (a \E[\ell_{t-1}] + b) \Big( (1-p_1)\dots(1-p_m) \nonumber \\
&+\frac{p_1(1-p_2)\dots(1-p_m)}{c_1\E[\ell_{t-1}] + d_1} + \dots + \frac{(1-p_1)(1-p_2)\dots p_m}{c_m\E[\ell_{t-1}] + d_m} \nonumber \\
&+ \frac{p_1p_2\dots(1-p_m)}{c_{12}\E[\ell_{t-1}] + d_{12}} + \dots + \frac{(1-p_1)\dots p_{m-1}p_m}{c_{m-1,m}\E[\ell_{t-1}] + d_{m-1,m}} \nonumber \\
&\vdots \nonumber\\
&+ \frac{p_1\dots p_m}{c_{1,\dots, m} \E[\ell_{t-1}] + d_{1,\dots, m}} \Big).
\end{align}
The final step now is to apply mathematical induction on $t$. Clearly the claim holds for $t=1$, since $\overline{\E[\ell_0]} = \E[\ell_0]$. Now assume that the claim holds for $t-1$, i.e., $\overline{\E[\ell_{t-1}]}$ is an upper bound on $\E[\ell_{t-1}]$ obtained as per the equality in Theorem~\ref{thm:stochastic}. At time $t$, we can write inequality~\eqref{eq:ell_ineq}. Now, the function $g(x) := (ax + b)/(cx+d)$, is monotonically increasing with $x \in \Real^+$ for any positive $c,d$ such that $ad > bc$ (This can be checked by verifying that $\partial g/\partial x > 0$). Therefore, each of the summing terms on the right hand side of~\eqref{eq:ell_ineq} are monotonically increasing functions of their argument, i.e., $\E[\ell_{t-1}]$. Therefore, substituting $\overline{\E[\ell_{t-1}]}$ in place of $\E[\ell_{t-1}]$, we obtain an upper bound on the right hand side, which is precisely the definition of $\overline{\E[\ell_t]}$. This completes the proof.
\end{IEEEproof}

\section{Proof of Corollary~\ref{thm:conservative_bound}}
\label{appx:proofTheorem3}

\begin{IEEEproof}[Proof of Corollary~\ref{thm:conservative_bound}]
We begin with the equality
\begin{align*}
&\E[\ell_{t+1} | l_t] = (a \ell_t + b)(1-p_1)\dots(1-p_m) \\
&+\frac{p_1(1-p_2)\dots(1-p_m)}{c_1\ell_t + d_1} + \dots + \frac{(1-p_1)(1-p_2)\dots p_m}{c_m\ell_t + d_m} \\
&+ \frac{p_1p_2\dots(1-p_m)}{c_{12}l_t + d_{12}} + \dots + \frac{(1-p_1)\dots p_{m-1}p_m}{c_{m-1,m}\ell_t + d_{m-1,m}} \\
&\vdots \\
&+ \frac{p_1\dots p_m}{c_{1,\dots, m} \ell_t + d_{1,\dots, m}}.
\end{align*}
We first collect all the terms that contain the term~$p_1$ in their numerator. For all of these terms, observe that the denominator terms can be lower bounded by~$c_1 \ell_t + d_1$, since each of the~$c$'s are at least equal to~$c_1$ and likewise for the~$d$'s. The numerator of those terms is each less than or equal to~$p_1$. Thus, we have
 \begin{align*}
&\E[\ell_{t+1} | \ell_t] \leq (a \ell_t + b)(1-p_1)\dots(1-p_m) +  \frac{p_1}{c_1\ell_t + d_1}\\
&+\frac{(1-p_1)p_2\dots(1-p_m)}{c_2\ell_t + d_1} + \dots + \frac{(1-p_1)(1-p_2)\dots p_m}{c_ml_t + d_m} \\
&+ \frac{(1-p_1)p_2 p_3 \dots(1-p_m)}{c_{23}\ell_t + d_{23}} + \dots + \frac{(1-p_1)\dots p_{m-1}p_m}{c_{m-1,m}\ell_t + d_{m-1,m}} \\
&\vdots
\end{align*}
Now, we repeat this procedure successively from the indices~$2$ through~$m$. Now, the function $g(x) := (ax + b)/(cx+d)$, is concave in $x \in \Real^+$ for any positive $c,d$ such that $ad > bc$ (This can be checked by computing $\partial^2 g/\partial x^2$). Therefore, from~\eqref{eq:cd}, each of the summing terms on the right hand side of the above equality are concave in the argument $\ell_{t-1}$. Unconditioning on the random variable $\ell_{t-1}$, and applying Jensen's inequality for each of the summing terms, we obtain
\begin{multline}\label{eq:cor}
\E[\ell_{t}] \leq (a \E[\ell_{t-1}] + b)  \times \\ \left(\prod_{j=1}^m(1-p_j)
+ \sum_{j = 1}^m \frac{p_j}{c_j\E[\ell_{t-1}] + d_j} \right).
\end{multline}
The final step now is to apply mathematical induction on $t$. Clearly the claim holds for $t=1$, since $\overline{\E[\ell_0]} = \E[\ell_0]$. Now assume that the claim holds for $t-1$, i.e., $\overline{\E[\ell_{t-1}]}$ is an upper bound on $\E[\ell_{t-1}]$ obtained as per the statement of Corollary~\ref{thm:conservative_bound}. At time $t$, we can write inequality~\eqref{eq:cor}. Now, the function $g(x) := (ax + b)/(cx+d)$, is monotonically increasing with $x \in \Real^+$ for any positive $c,d$ such that $ad > bc$ (This can be checked by verifying that $\partial g/\partial x > 0$). Therefore, each of the summing terms on the right hand side of~\eqref{eq:ell_ineq} are monotonically increasing functions of their argument, i.e., $\E[\ell_{t-1}]$. Therefore, substituting $\overline{\E[\ell_{t-1}]}$ in place of $\E[\ell_{t-1}]$, we obtain an upper bound on the right hand side, which is precisely the definition of $\overline{\E[\ell_t]}$. This completes the proof.
\end{IEEEproof}
\end{document}